# Self-imitating Feedback Generation Using GAN for Computer-Assisted Pronunciation Training

*Seung Hee Yang[1], Minhwa Chung[1, 2]*

[1]Interdisciplinary Program in Cognitive Science, Seoul National University, Republic of Korea
[2]Department of Linguistics, Seoul National University, Republic of Korea
sy2358@snu.ac.kr, mchung@snu.ac.kr

## Abstract

Self-imitating feedback is an effective and learner-friendly method for non-native learners in Computer-Assisted Pronunciation Training. Acoustic characteristics in native utterances are extracted and transplanted onto learner's own speech input, and given back to the learner as a corrective feedback. Previous works focused on speech conversion using prosodic transplantation techniques based on PSOLA algorithm. Motivated by the visual differences found in spectrograms of native and non-native speeches, we investigated applying GAN to generate self-imitating feedback by utilizing generator's ability through adversarial training. Because this mapping is highly under-constrained, we also adopt cycle consistency loss to encourage the output to preserve the global structure, which is shared by native and non-native utterances. Trained on 97,200 spectrogram images of short utterances produced by native and non-native speakers of Korean, the generator is able to successfully transform the non-native spectrogram input to a spectrogram with properties of self-imitating feedback. Furthermore, the transformed spectrogram shows segmental corrections that cannot be obtained by prosodic transplantation. Perceptual test comparing the self-imitating and correcting abilities of our method with the baseline PSOLA method shows that the generative approach with cycle consistency loss is promising.

## 1. Introduction

Generating a corrective feedback is an important issue in the area of spoken language technology for education [1]. This process may be painstaking if each corrected utterance has to be recorded by a teacher, or result in a negative outcome if the automatic generation of the ideal form is not correctly synthesized. Studies on computer assisted pronunciation training (CAPT) found that the better the match between the learners' and native speakers' voices, the more positive the impact on pronunciation training [2,3]. This emphasizes the importance of the student and teacher voice similarity for the enhancement of pronunciation skills. In self-imitating feedback, the characteristics in native utterances are extracted and transplanted onto the learner's speech. Listening to the manipulated speech enables students to understand the differences between their accented utterances and the native counterparts, and to produce native-accented utterances by self-imitation.

In previous works, speech conversion methods for pronunciation teaching have been studied for Korean and Japanese learners of English, Italian learners of German, Japanese learners of Italian, and for English learners of Mandarin Chinese [4,5,6,7,8]. These studies were based on the prosodic transplantation technique [9], using PSOLA (Pitch-Synchronous Overlap and Add) algorithm [10]. Through this technique, the acoustic parameters including pitch, intensity, articulation rate, and duration of the native speakers are transferred to the learners' speech.

These studies have shown that corrective feedback can be successfully generated at the suprasegmental level. However, the proficiency in a second language is fully attained only if the students have learned to modulate both the prosodic and segmental parameters equivalent to those of the native speakers. The previous methods have been limited to the prosodic level only, although the segmental accuracy plays an important role in spoken language communication [11].

In the first part of this work, we conduct a linguistic comparison between native and non-native utterances by visualizing their differences in pairs of spectrograms, i.e., time-frequency representations of speech. The spectrogram analyses illustrate the segmental characteristics between the two domains, which motivates our idea for using image-generating generative adversarial network (GAN) [12] to learn the mappings between native and non-native spectrograms. Our approach is a first attempt, to the best of our knowledge, at using GAN for speech correction. Since there are numerous "golden references" by native speakers, there are infinitely many mappings the generator can learn. Assuming that there is some underlying relationship between non-native and native linguistic domains, we further adopt cycle consistency loss [13] to induce the output to preserve the global structure, which is shared by native and non-native utterances. We then compare the proposed method with the baseline PSOLA method through a perceptual evaluation, during which corrective and self-imitative effects are judged by human experts.

There are potential advantages of the proposed GAN-based corrective feedback generation. First, GANs enable simplified procedure in feedback generation because it does not rely on the intermediate processes of feature extraction and error region detection. Second, translating spectrograms for sound manipulation in language learning is immediately useful, such as in CAPT applications. Third, the discriminator in GAN has the ability to judge the nativelikeness of spectrograms, which can be used to perform speech assessment task in language learning [14]. Furthermore, despite their increasing fidelity at translating static images [15,16], GANs have yet to be demonstrated to be capable of translating spectral representations of audio, which is the main issue of this paper.

## 2. Linguistic Differences between Native and Non-native Speech

One way to analyze speech is by examining their spectrograms, which visually represent the varying short term amplitude spectra of the speech waveform. Spectrogram analysis contains the information on phonetic characteristics, and the practice of using them for speech recognition tasks is common in the discriminative setting [17]. We first make observations on the differences between native and non-native speech by comparing spectrogram pairs of the utterances for the same words in Korean.

In Figure 1, we show an example of a spectrogram pair for the word "half a year." While the left spectrogram captures the resonances of the vocal tract during a diphthong articulation, the right spectrogram shows its monophthong version. As a consequence, the two spectrograms can be differentiated by the number and movements of the darkness bands, showing that non-native speeches are more likely to substitute diphthongs by monophthongs than the native speech. By observing more spectrogram examples, we obtain linguistic differences including final stop deletions, exhibited by the voiced and unvoiced region contrasts in the spectrograms, and lenition of tense consonants, which is demonstrated by the voice onset time in the spectrograms. Moreover, the presence of rhotic vowels in the formant frequencies of the non-native spectrograms is not observed for native counterpart, as the sound does not exist in its phonetic inventory. At the suprasegmental level, the articulation rate and total duration of the native speakers tend to be shorter than the learners' speech. These findings can be confirmed by analyses of the auditory variation patterns in [18].

Based on these observations, we draw two implications for corrective feedback generation. First, we find that spectrograms contain rich information that is enough to differentiate the characteristics of native and non-native utterances in linguistic domains. This motivates our idea for a spectrogram learning using image-generating GAN, where latent space in the audio of non-native linguistic domain is mapped to that of native linguistic domain. Second, despite the differences between the two domains, we also find that they share an underlying structure. Different renderings of the same speech are possible since there can be numerous "golden references," and in theory, there are infinitely many possible acceptable outputs. In order to avoid such confusion, it seems desirable that the outputs preserve the global structure in the input spectrograms. In the following section, we explore how GAN-based methods can exploit these properties.

## 3. Feedback Generation using GAN

### 3.1. Generative Adversarial Networks

GANs have attracted attention for their ability to generate convincing images and speeches. GANs [12] are generative models that learn to map the training samples to samples with a prior distribution. The generator (G) performs this mapping by imitating the real data distribution to generate fake samples. G learns the mapping by means of an adversarial training, where the discriminator (D) classifies whether the input is a fake G sample generated by G or a real sample. The task for D is to correctly identify the real samples as real, and thereby distinguishing them from the fake samples. The adversarial characteristic is due to the fact that G has to imitate better samples in order to make D misclassify them as real samples.

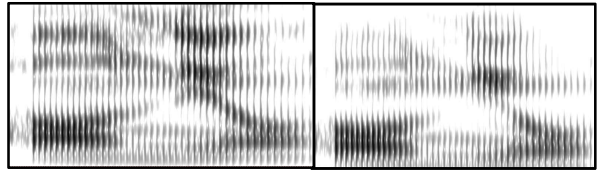

Figure 1. *An example of a spectrogram pair for the word "half a year (반년)" in Korean uttered by a native (left) and non-native (right) speakers. We observe that spectrogram comparisons are able to capture linguistic differences, which motivates our corrective feedback design choices.*

The misclassification loss is used for further improvement of the generator. During the training process, D back-propagates fake samples from G and correctly classifies them as fake, and in turn, G tries to generate better imitations by adapting its parameters towards the real data distribution in the training data. In this way, D transmits information to G on what is real and what is fake. This adversarial learning process is formulated as a minimax game between G and D, which is formulated as:

$$\min_G \max_D V(D,G) = \mathbb{E}_{x \sim P_{data}(x)}[\log D(x)] + \mathbb{E}_{z \sim P_z(z)}[\log(1 - D(G(z)))]. \quad (1)$$

where $P_{data}(x)$ is the real data distribution, and $P_z(z)$ is the prior distribution. For a given x, $D(X)$ is the probability x is drawn from $P_{data}(x)$, and $D(G(z))$ is the probability that the generated distribution is drawn from $P_z(z)$. analyses of the auditory variation patterns in [18].

Conditional GANs (cGANs) learn a conditional generative model [19] where we condition on the input and generate corresponding output. G tries to minimize the objective below against an adversarial D that tries to maximize it.

$$L_{cGAN}(G, D) = \mathbb{E}_{x,y}[\log D(x,y)] + \mathbb{E}_{x,y}[\log(1 - D(x, G(x,z)))]. \quad (2)$$

[19] demonstrated that cGANs can solve a wide variety of problems by testing the method on nine different graphics and vision tasks, such as style transfer and product photo generation. By interpreting speech correction task as a spectrogram translation problem, we explore the generality of conditional GANs.

With large enough capacity, the adversarial loss alone may not guarantee that the learned function can map the input to the desired output. In our case, this may result in inappropriate or unwanted corrections generated by the network, which is highly undesirable for self-imitating learning. [13] introduced cycle consistency loss to further reduce the space of possible mapping functions. This is incentivized by the idea that the learned mapping should be cycle-consistent, which is trained by the forward and backward cycle-consistency losses:

$$L_{cyc}(G, F) = \mathbb{E}_{x \sim P_{data}(x)}[\|F(G(x) - x\|_1] + \mathbb{E}_{y \sim P_{data}(y)}[\|G(F(y) - y\|_1]. \quad (3)$$

Here, the network contains two mapping functions G : X → Y and F : Y → X. For each image *x* from domain X, the translation cycle should be able to bring *x* back to the original image, and vice versa. While the adversarial loss trains to match the distribution of generated images to the data distribution in the target domain, the cycle consistency losses can prevent the learned mappings G and F from contradicting each other. In addition to the conditional GAN, we explore the generator's behavior when trained with the full objective including adversarial and cycle consistency losses.

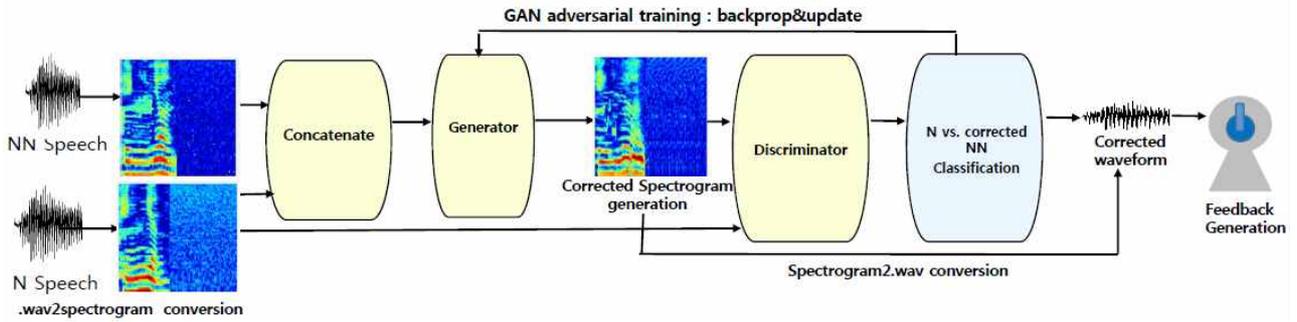

Figure 2. *Experiment Method using the speech2wav conversion, spectrogram learning using conditional GAN, and wav2speech conversion for feedback generation*

### 3.2. Self-Imitating Feedback Generation using GAN

The proposed method using GAN is done in five steps: 1) native (N) and non-native (NN) paired speech preparation, 2) speech-to-spectrogram conversion, 3) spectrogram-to-spectrogram training, 4) inversion back into audio signal, and 5) playback the generated audio back to the learner. GAN is used in the third step and the conversion techniques are used during the second and the fourth steps. In order to train using conditional GAN, the prepared samples are first concatenated and fed into the generator, where adversarial training is done using the discriminator which classifies whether the samples are fake (generated/corrected speech) or real (native speech). The process is shown in Figure 2. For the cycle-consistent adversarial training, there is no concatenation step, since it takes unpaired input.

## 4. Experimental Method

### 4.1. Corpus

The proposed model is trained on L2KSC (L2 as Korean Speech Corpus) [20]. The corpus is used because it is a parallel native and non-native speech database available to the public and fits our experiment settings. There are 217 non-native speakers with 27 mother tongue backgrounds, and 107 native speakers of 54 females and 53 males. Each speaker read 300 short utterances, which are in average one second in length. When each spectrogram of non-native recording is paired with all native recordings of the same utterance, there are 1,357,321 pairs of samples for the conditional GAN training. For cycle consistent adversarial training, there are 32,100 and 65,100 spectrograms in the native and non-native domains, each respectively. The 162 spectrograms for test are completely held-out.

### 4.2. Experiments

#### 4.2.1. Baseline Implementation

Baseline corrective feedback sounds were generated using PSOLA algorithm, implemented in Praat [21]. The acoustic parameters of pitch, intensity, and duration of the native speech of the same utterance are extracted and transplanted on to the held-out non-native recordings manually to provide the best performance of PSOLA algorithm.

#### 4.2.2. Speech-to-Spectrogram and Spectrogram-to-speech Conversions

We first convert audio signal to spectrogram using Short-Time Fourier Transform (STFT) with windows of 512 frames and 33% overlap, converted to dB amplitude scale, represented using mel scale and padded with white noise to generate 128x128 pixels images.

We use griffin_lim framework [22] which is a python implementation of the Griffin and Lim algorithm to convert the spectrogram to audio signal by using the magnitude of its STFT. It performs low-pass filtering of the spectrogram by zeroing all frequency bins above the preset cutoff frequency, and then uses the Griffin and Lim algorithm to reconstruct an audio signal from the spectrogram. The algorithm works to rebuild the signal with STFT such that the magnitude part is as close as possible to the spectrogram. For high quality output and minimum loss in transformations, it is run for 1,000 iterations. Perceptual evaluation of the regenerated audio signal before and after transformation do not show any significant difference in quality. The different utility tools built around the framework are released on our github repository.

#### 4.2.3. Spectrogram2Spectrogrm Training

The spectrogram2spectrogrm translation for conditional GAN follows the same network architecture as in Pix2Pix framework [19], which uses "U-Net" shaped generator [23] with skip connections that allows to capture low-level information shared by the input and output while circumventing the information loss at the bottleneck. For the discriminator training, Markovian PatchGAN [19] is used for classifying if each N x N patch in an image is real or fake. The CycleGAN framework [13] is used for the cycle consistent adversarial training, which adopts the generator architectures from [24], which has shown impressive results for neural style transfer, and uses PatchGAN discriminator.

Native and non-native pairs of spectrograms corresponding to the same utterances are taken as the input into the Pix2Pix framework, while unpaired 97,200 spectrogram images of the two domains are fed into the CycleGAN network. Data augmentation option by flipping images is disabled and batch size was increased to 4 from the default 1. When the training is finished, the model is applied to all the test spectrograms. Web interface visualization of the training process, which was offered in the frameworks, was used to monitor the training and track how spectrogram and the corresponding sounds evolve over time. The supplementary material, "Training Process Visualization with Sound.mp4," shows a case of corrective evolution for the word "first time (처음)," where the spectrogram is correcting the syllable deletion error.

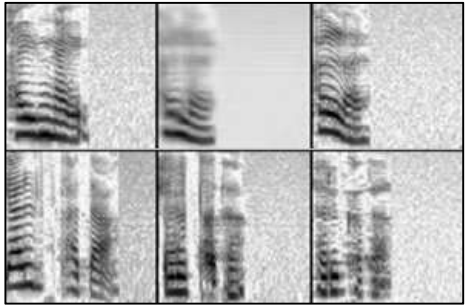

Figure 3. *Spectrogram learning using Pix2Pix framework for non-native, generated, and native speeches, from left to right, at epoch 1 (above) and epoch 3 (below).*

## 5. Results and Evaluation

### 5.1. Spectrogram Generation Results

Figure 3 shows the spectrograms for non-native, generated, and native speeches at epoch 1 and epoch 3 in the Pix2Pix framework. It shows that the generator quickly learns to imitate the native spectrogram by generating a fake version of the reference. After more training, the generator has discovered to generate spectrograms with higher proximity to the native. Since the test data was completely held out, this means that the model learned to recognize which word the spectrogram represents, and identified which native spectrogram should be mapped to the given non-native.

### 5.2. Perceptual Evaluation

#### 5.2.1. Method

Our ultimate goal is to produce examples that are corrective, self-imitative, and intelligible to humans. To this end, we measure the ability of human annotators to label the generated audio. Using our three models, PSOLA, pix2pix, and CycleGAN, we generate evaluation files, which amount to 486 waveforms in total. Examples of the spectrograms before the conversion are shown in Figure 4. The four native Korean human raters with knowledge in linguistics assigned subjective values from 1 to 5 for the five criteria: holistic impression of correction, degree of segmental correction, degree of suprasegmental correction, sound quality, and speaker voice imitability. The score of 3 was assigned if there is no difference before and after the manipulation. They listened to the original non-native utterance, followed by a generated output from one of the three models. The order of presentation was randomized.

#### 5.2.2. Result

We report MOS (mean opinion scores) values in Table 1. It shows that our newly proposed CycleGAN-based speech correction method is able to generate corrective feedback. Linguistic analysis shows that the generator's corrective ability is effective both in the segmental and suprasegmental aspects. Since an error in the generated feedback can be critical in learning applications, we verified that all corrective ability scores in CycleGAN are 3 or above, which means that there was no degradation. For the baseline PSOLA method, the evaluators report that there were numerous cases when the generated results does not make corrections, or make corrections that are perceptually trivial. On the other hand, the generated results using Pix2Pix framework often fails to generate a corrected speech. The supplementary material, "Test Data Visualization with Sound.mp4," enables direct comparisons with auditory data.

Table 1: *MOS values of perceptual test by four human experts on self-imitation feedback generation (SQ: Sound Quality)*

| Model | Corrective Ability | | | Imit-ability | SQ | Avg. |
|---|---|---|---|---|---|---|
| | Holis-tic | Seg-mental | Supra-segmental | | | |
| PSOLA | 3.118 | 3.029 | 3.324 | 4.029 | 2.794 | 3.259 |
| Pix2Pix | 1.970 | 2.485 | 2.152 | 2.697 | 1.636 | 2.188 |
| Cycle-GAN | 4.000 | 4.333 | 4.364 | 3.515 | 2.667 | 3.776 |

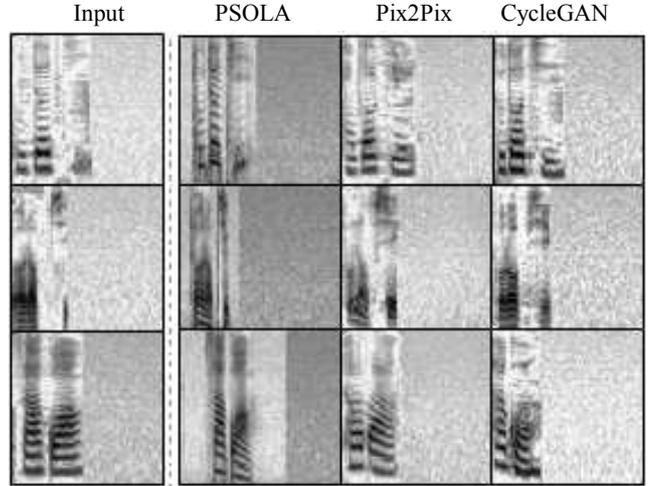

Figure 4. *Different methods for corrective generation by using PSOLA algorithm and spectrogram learning.*

In addition to MOS scores, we conducted auditory transcription of the generated utterance on a random subsample of the test set for qualitatively analyzing where the correction occurs. Successful cases include corrections of detensifying errors of /s͈/ in the word "fishing (낚시)," as mentioned in the spectrogram comparisons in Section 2. Moreover, while the statement "It is fast (빨라요)" was realized as a question with a final rise, it was corrected by the generator. The rate of speech tends to be closer to the native when there were silence between syllables in the non-native speech. We also found cases of negative correction, such as omitting a syllable or a final stop.

In all cases, the generated sound qualities were worse than the original recording. For the two generative models, it possible that the poor qualitative ratings are primarily caused by the lossy Griffin-Lim inversion, therefore, synthesizing clear audio needs to be addressed in the future work. Moreover, there is a room for improvement in CycleGAN's imitability score, which is lower than PSOLA method. This may be due to the diversity in reference styles and future work can be expanded for the generator to better imitate speaker voice characteristics.

## 6. Conclusion

This study lays the groundwork for an automatic self-imitating speech correction system for pronunciation training. To the best of our knowledge, it is the first approach comparing different GAN architectures on spectrogram. The perceptual evaluation shows that cycle consistent adversarial training is a promising approach for speech correction task. In our future work we plan to extend to improve speaker voice imitability and operate on longer length audio recordings to explore a variety of conditioning strategies.